\documentclass[
]{ceurart}

\usepackage{multirow}
\usepackage{url}
\usepackage{tabularx}
\usepackage{makecell}
\usepackage{graphicx}

\begin{document}

\copyrightyear{2023}
\copyrightclause{Copyright for this paper by its authors.
  Use permitted under Creative Commons License Attribution 4.0
  International (CC BY 4.0).}

\conference{}

\title{How to Evaluate Coreference in Literary Texts?}

\author{Ana-Isabel Duron-Tejedor}
\author{Pascal Amsili}
\author{Thierry Poibeau}
\address{Lattice, CNRS \& ENS-PSL \& Université Sorbonne Nouvelle}


\begin{abstract}
  In this short paper, we examine the main metrics used to evaluate textual coreference and we detail some of their limitations. We show that a unique score cannot represent the full complexity of the problem at stake, and is thus uninformative, or even misleading. We propose  a new way of evaluating coreference, taking into account the context (in our case, the analysis of fictions, esp. novels). More specifically, we propose to distinguish long coreference chains (corresponding to main characters), from short ones (corresponding to secondary characters), and singletons (isolated elements). This way, we hope to get more interpretable and thus more informative results through evaluation.  \end{abstract}

\begin{keywords}
  Coreference \sep
  Evaluation \sep
  Metrics \sep
  Literature analysis
\end{keywords}

\maketitle

\section{Introduction}

Coreference resolution is a natural language processing (NLP) task that consists in determining the linguistic expressions, also called mentions, that refer to a same entity \citep{poesio2023}. This task is crucial for NLP, and constitutes a logical complement to named entity recognition. For example, if one wants to track all the actions involving a given entity in a text, an accurate analysis of coreference is required, as most of the references to entities are usually done through coreference (i.e.,  pronouns and other linguistic items referring to an entity). 

This is in particular the case for the analysis of fictions, especially novels: a proper coreference analysis is mandatory to analyze the plot, the evolution, and the interactions between the different characters \citep{bamman2014}. 
Characters play a prominent roles in fiction. For this reason, we will only consider reference to characters in this study (which means we exclude, e.g., locations and company names, that are of course useful, but do not play a so prominent role in fictions). 

The importance of coreference in NLP has been put forward by the Message Understanding Conferences (MUC) in the late 90ies \citep{grishman1996}. Since then, numerous coreference resolvers have been developed for a variety of languages and coreference is a very active domain of research in NLP. 
A related issue is how to evaluate coreference. As coreference involves sets of mentions of various size linked together (i.e., referring to a same entity), traditional indicators like precision and recall have to be redefined, but there is no obvious way of doing it (should entities be weighted based on their relative importance in the text? What to do with singletons?). This led to several measures that do not always agree (i.e. one scoring method may indicate a good precision, whereas another method will not be so positive). 

This has been already been demonstrated in various studies on the topic \citep{borovikova2022,lion-bouton2020,moosavi2016,moosavi2020}. Here we want to stress complementary, albeit different, issues, and more specifically two of them:
\begin{itemize}
    \item Existing systems and evaluation methods are based on small texts, and are not fully relevant when addressing long texts like novels;
    \item Existing evaluation methods tend to consider all the entities on an equal basis, regardless of the linguistic properties of the entity,  which is not the case for most practical applications. 
\end{itemize}    
We will also point out a more general issue: the fact that none of the existing evaluation methods is really interpretable by a human. 

The paper is structured as follows: we first remind the reader with state-of-the art metrics and the main issues that have been reported about them. We then show that these measures have been defined for small texts but long texts like fictions tend to exacerbate the issues detailed in the previous section. We finish with some proposals based on a typology of coreference chains, to get more readable and actionable results. 

\section{State of the Art in Coreference Evaluation Methods}

Coreference evaluation metrics can be divided in two categories: link-based and mention-based metrics. Roughly, link-based metrics represent entities as a set of links and mention-based represent them as a set of mentions. In this section, we just give a brief overview of the most popular methods and of their relative weaknesses, as more detailed overviews already exist. The interested reader can refer to \citep{borovikova2022,lion-bouton2020,moosavi2016,moosavi2020}. Traditionally, in the literature on coreference, `key' refers to the reference (gold annotations), and `response' corresponds to the computed links. 

\textit{MUC} \citep{vilain1995} is only based on the minimum number of missing/extra links in the response compared to the key entities. It does not take into account singletons, nor the relative weight of the different entities (all the links have the same weight). 

\textit{BLANC} \citep{recasens2011} is a metric that was built in a first place to evaluate coreference on key mentions. It was then adapted \citep{luo2014} 
to also evaluate response mentions. Contrary to MUC, BLANC uses non-coreferent links. The major problem with BLANC is that, when the number of gold mentions in the response entities is larger, the number of non-coreference links is also larger. In this case BLANC gives higher scores (for both recall and precision) without knowing if the gold mentions were resolved \citep{moosavi2016}. 

Two other measures, $B^3$ and \textit{CEAF}, are mention-based. In the case of $B^3$ \citep{bagga1998}, the overall recall/precision is computed based on the recall/precision of the individual mentions. For each mention m in the key entities, $B^3$
recall considers the fraction of the correct mentions that are included in the response entity of m. Similarly to MUC, $B^3$ precision is computed by switching the role of the key and response entities \citep{moosavi2020}.

\textit{CEAF} \citep{luo2005} relies on a one-to-one mapping, where each key entity is mapped to only one response entity (\textit{CEAF}$_e$ is based on the mapping at the entity level, a variant \textit{CEAF}$_m$ works at the mention level). Via this mapping, \textit{CEAF} uses a similarity metric $\phi$(i,j) to measure the similarity of the mapped entities with the reference. The one-to-one mapping at the heart of \textit{CEAF} leads to ignoring other correct decisions of unaligned
response entities. Finally, \textit{CEAF}$_e$ does not assign weights to the different entities: it can be seen as a metric that gives a global score of the proportion of entities that were found in the response (for  recall) and the proportion of correct entities in the response (for precision).

All these metrics have been studied, compared and criticized, esp. in  \citep{moosavi2016,moosavi2020}. They suffer from a number of known issues, especially what \citep{moosavi2016} call the ``mention identification effect''. The authors propose the following experiment: they remove spurious mentions from the responses entities (the ones calculated automatically). This should automatically increase precision (since there are then less spurious mentions), while having no effect on recall. Surprisingly, the authors observe the following, ``(\textit{i}) recall changes for all the metrics except for MUC; (\textit{ii}) both $CEAF_e$ recall and
precision significantly decrease; and (\textit{iii}) BLANC
recall notably decreases so that F1 drops significantly in comparison to Base'' \citep{moosavi2016}. 

As a consequence, they propose \textit{LEA}, a new method supposed to overcome the previous limitations. 
LEA \citep{moosavi2016,moosavi2020} represents the importance of each entity through a weight (meaning that LEA considers that a spurious singleton is less harmful than
a spurious long entity). Since LEA is link-based, the metric is not affected by the
mention-identification effect \citep{moosavi2016}, which causes mention-based metrics scores to be less interpretative. 
Finally, even if LEA gives a representation to singletons (i.e. a self-link), a singleton is considered to be found only if the singleton is correctly resolved. 

Despite these improvements, it has been demonstrated that LEA also suffers from a number of problems \citep{lion-bouton2020}. More specifically, LEA does not support specific configurations (esp. what the authors of \citep{lion-bouton2020} call `triangular inequality', which is not a rare case). Very similar configurations may lead to very different scores, which means systems cannot be compared and results cannot be directly interpreted. 

\section{The Inadequacy of Traditional Evaluation Schemes for Literary texts}

In this section, we observe the application of the previous measures (that constitute de facto standard, despite their limitations) to the analysis of long texts, especially fictions and novels.  

\subsection{The Inadequacy of Reference Corpora}

Coreference is usually evaluated on a specific corpus called OntoNotes. OntoNotes was built in the 2000s by a group of researchers from different American universities and covering a number of structural annotations \citep{hovy2006}. This corpus has been especially popular for coreference evaluation, as it is one of the very rare resources available for the task. It gathers different kinds of texts (although they are mostly news) and is thus considered quite balanced and representative. 

However, the OntoNotes corpus suffers from a number of shortcomings. The corpus is made of short texts (including short extracts of longer texts), which means coreference chains are local and limited in size. The task is thus limited in its complexity, for example the set of antecedents for a given pronoun is small, given the short size of the text to consider. Coreference is also centered around pronouns, whereas real texts provide a wider diversity of coreferences, especially definite descriptions (''the lady'', ''the milkman''). 

Long texts, especially novels, are of course very different. Table~\ref{tabComparison} reports some numbers that show  a clear difference between OntoNotes and a novel  (\textit{Manon Lescaut} in this case --- numbers would be different with other novels but the same tendency could be observed).

\begin{table}[ht]
\begin{tabular}{|l|r|r|}
    \hline
  \makegapedcells
     &  \textbf{Ontonotes ~ ~ ~} & \textbf{Manon Lescaut} \\
     &  (average over ~ ~  &  \\
     &  3,493 documents) &  \\
    \hline
    \# of mentions & 56 ~ & 11,888 ~ \\
    \hline
    \# of chains & 13 ~ & 143 ~ \\
    \hline
    \# of mentions per Chains (incl. singletons) & - ~ & 59 ~ \\
    \hline
   \# of mentions per Chains (without singleton) & 4.1 ~ & 83 ~ \\
    \hline
    \# of singletons  & 0 ~ & 56 ~  \\
    \hline
    \# of tokens & 466 ~ & 77,232 ~ \\
    \hline
\end{tabular}
\label{demo-table}
\caption{Comparison of Ontonotes with a typical French novel, \textit{Manon Lescaut}.}
\label{tabComparison}
\end{table}

As one can see, the differences are huge. While OntoNotes documents are made of a few short chains (4.1 mentions per chain on average), novels like \textit{Manon Lescaut} have 83 mentions per coreference chain on average. Moreover the distribution of chain lengths is very skewed, as nearly half of the mentions in  Manon Lescaut correspond to one character only (the so called \textit{Chevalier Des Grieux}), see Figure 1. 

\begin{center}
\includegraphics[scale=.3]{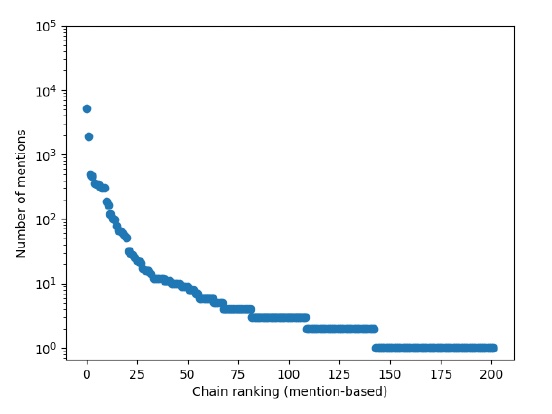}
\captionof{figure}{The distribution of coreference chains in \textit{Manon Lescaut}, based on the number of mentions per chain. As one can see, the distribution is not even Zipfian (in which case, the curve would be straight), the number of very long chains being extremely limited. Experiments with other novels, by Balzac among others, show a quasi Zipfian curve, showing that more characters with first and secondary roles appear in Balzac's world.}
\label{fig1}
\end{center}

Even if \textit{Manon Lecaut} is thus an extreme case, most novels are centered around a handful of characters that correspond to most of the mentions. Lastly, one may notice differences due to the annotation schemes, as OntoNotes does not consider singletons for the coreference task.
As a conclusion, we think that even if OntoNotes is a \textit{de facto} standard for coreference evaluation, it cannot be used as an indicator of future performance on the fiction domain.

\subsection{The Low Readability of Existing Evaluation Methods}

Besides the known limitations of existing evaluation methods that we have seen in section 2, these metrics are not ideal in that they all give different scores for a same input (but these differences are not directly interpretable). 
For example, Pradhan et al. \citep{pradhan2014} describe precisely how these metrics work on a toy example. In the end, their example corresponds to the following scores (F-measure): 0.4 for MUC, 0.46 for B3, 0.52 for CEAF and 0.36 for Blanc (so 0.16 difference between Blanc and CEAF). 
Given that different choices can be made concerning the way to calculate precision and recall (mention-based or relation-based, taking into account chain length or not, etc.) it is normal that each method has different scores for precision and recall. However, this remains a major issue. 

The problem is exacerbated in the case of long novels, as one can imagine. Other issues concern the reference: some corpora/experiments consider only pronouns, some others also definite descriptions (like ``the coachman''), which should be part of the task but are harder to annotate and evaluate. 
For example, Bamman et al. \citep{bamman2020} say that ``OntoNotes and LitBank do not link two generic common nouns, while PreCo does'' (OntoNotes, LitBank and PreCo are three corpora used in Bamman et al.'s paper), which makes the comparison between the different results delicate. The annotation of singletons also differs a lot from one corpus to the other. All these are known problems, that have reinforced the status of OntoNotes as a de facto standard  (since OntoNotes provides stability, although the corpus is limited and not always relevant). 



Given all these differences, a known strategy consists in calculating the mean of the different evaluation scores (or of a subset of these scores -- for example \citep{denis2009} propose to combine MUC, CEAF and B3), a score sometimes called the CoNLL score (since it was used at CoNLL 2012 \citep{pradhan2012}). In our opinion, this is probably worse than keeping the different scores apart: since these scores have few in common, calculating the mean does not really make any sense (like calculating the mean of an heterogeneous set of elements). 

These scores may be useful in an NLP perspective, if one really needs or wants to compare systems (although we, and many others, have shown that this comparison is dubious). But they are probably useless in a Digital Humanities perspective (or just from a linguistic point of view), where one needs to have a clear view of what is going on. In this context, the key question is to know what kinds of coreference chains are accurately analyzed, and which ones may suffer from errors. 

\section{Proposals for a Better Evaluation of Coreference in Long Documents}

Following what we have shown in the previous sections, we think that a unique score is unable to take into account all the complexity of coreference. Moreover, a unique score (or even a combination of scores) is not really useful in the context of fictions, since no score is really interpretable in itself and scores are not comparable with each other. 

It would thus make more sense to promote indicators based on the linguistic content that is measured. In the case of fictions for example, at least three different cases should be considered:

\begin{itemize}
    \item long coreference chains, especially those including a named entity (i.e. where one of the mentions is a named entity). This generally corresponds to main characters and is especially important for most analyses (esp. representing the plot). 
    \item singletons. We have seen that most metrics have a hard time with singletons, since by definition a singleton (a single mention) is not a chain (that requires at least two mentions linked together). Hence the choice of MUC (that ignores singletons) or LEA (that consider singleton as an entity with a link to itself). Singletons are important to evaluate the ``population in a novel'' (the number of characters in a novel) but should be considered in isolation and probably not as a  coreference element (of course, a singleton can be included in a longer coreference chain by mistake, but in such a case, it would just be a mistake that is independent from the singleton issue). 
    \item remaining coreference chains refer for the most part to secondary characters or just `background characters', who are generally not named but can be important for the plot. Being able to evaluate the accuracy of the analysis on this kind of character is thus key. It can also be problematic, as the issue is not always fully decidable. For example, if there are different mentions of `coachman' in a novel (referred to just as 'coachman'), should we consider that they all refer to the same character, or not? This kind of precision is probably not possible in a number of cases in each novel, which means that calculating precisely the ``population of a novel'' is not possible with perfect accuracy \citep{woloch2004,lavocat2020}.  
\end{itemize}

Traditional scores can then be calculated for the different cases and thus give a better picture of the situation at stake. This method may make it more difficult to compare different systems, but existing measures do not allow such a comparison anyway. A more fine-grained analysis based on the linguistic content would probably be more useful for digital humanists than current NLP-based way of reporting results. 

For example, when analyzing the BookNLP\footnote{\url{https://www.lattice.cnrs.fr/projets/booknlp/}} scores on the French LitBank\footnote{\url{https://github.com/lattice-8094/fr-litbank}}, it was observed that long coreference chains were sometimes broken in different parts, thus showing the necessity to reinforce the clustering dimension of the algorithm (this procedure makes it possible to calculate coreference over different mentions of a same entity, thus reinforcing the quality of the analysis despite surface variation). 

The same strategy can be used to track the analysis of singletons for example. This is in line with the approach described in \citep{bamman2020}, where the authors propose some manipulations and an evaluation with the $B^3$ metric to study precisely the treatment of singletons with their approach (section 6.1, page 51 of \citep{bamman2020}). The same approach can be done on the French LitBank to refine this part of the analysis.

\section{Conclusion}

In this short paper, we presented known issues with evaluation scores related to coreference, and we have shown that this inadequacy is especially meaningful for long texts. We have proposed a new way of evaluating coreference, taking into account the application context (digital humanities) as well as the need to have more informative and interpretable metrics. In the future, we will apply this approach to more novels, in different languages, to ensure the approach gives an appropriate representation of the problem, help the user spot weaknesses in her/his models and, above everything, ensure that the metrics used are both readable and interpretable.

\bibliography{bibliography}


\end{document}